\documentclass{article}

\usepackage{microtype}
\usepackage{dsfont}
\usepackage{graphicx}
\usepackage{multirow}
\usepackage{subcaption}
\usepackage{amsmath,amsfonts,bm,bbm}

\usepackage{booktabs} 
\usepackage[disable]{todonotes}

\usepackage[draft=false]{hyperref}  

\usepackage{makecell, longtable}

\usepackage{etoolbox, siunitx, booktabs}

\newcommand{\numobjective}{L}
\newcommand{\numdataset}{M}
\newcommand{\numparameters}{p}
\newcommand{\numevaluations}{N}
\newcommand{\mlpparameters}{\theta}
\newcommand{\ECDF}{\tilde{F}}
\newcommand{\priormu}{\mu_\mlpparameters}
\newcommand{\priorsigma}{\sigma_\mlpparameters}
\newcommand{\surrogate}[1]{\hat{#1}}
\newcommand{\standardize}{\Delta}

\newcommand{\task}[1]{{#1}}
\newcommand{\DeepAR}{\task{DeepAR}{}}
\newcommand{\FCNET}{\task{FCNET}{}}
\newcommand{\NAS}{\task{NAS}{}}
\newcommand{\XGBoost}{\task{XGBoost}{}}

\newcommand{\baseline}[1]{{#1}}
\newcommand{\RS}{\baseline{RS}{}}
\newcommand{\GP}{\baseline{GP}{}}
\newcommand{\ABLR}{\baseline{ABLR}{}}
\newcommand{\WSGP}{\baseline{WS-GP}{}}
\newcommand{\SGPT}{\baseline{SGPT}{}}
\newcommand{\AutoGP}{\baseline{AutoGP}{}}

\newcommand{\CTS}{\baseline{CTS}{}}
\newcommand{\TS}{\baseline{TS}{}}
\newcommand{\GCP}{\baseline{GCP}{}}

\newcommand{\EHI}{\baseline{EHI}{}}
\newcommand{\EMI}{\baseline{EMI}{}}
\newcommand{\SUR}{\baseline{SUR}{}}
\newcommand{\SMS}{\baseline{SMS}{}}

\newcommand{\GPprior}{\baseline{GP+prior}{}}
\newcommand{\GCPprior}{\baseline{GCP+prior}{}}

\newcommand{\ADTM}{\text{ADTM}}

\newcommand{\DTM}{\text{DTM}}

\newcommand{\N}{\mathcal{N}}

\newcommand{\R}{\mathbb{R}}
\newcommand{\E}{\mathbf{E}}

\DeclareMathOperator*{\argmax}{arg\,max}
\DeclareMathOperator*{\argmin}{arg\,min}

\usepackage[accepted]{icml2020}

\usepackage{import}
\graphicspath{{figures/}}

\usepackage{xr}
\begin{document}
\robustify\bfseries
\twocolumn[
\icmltitle{A Quantile-based Approach for Hyperparameter Transfer Learning}




\begin{icmlauthorlist}
\icmlauthor{David Salinas}{A}
\icmlauthor{Huibin Shen}{B}
\icmlauthor{Valerio Perrone}{B}
\end{icmlauthorlist}

\icmlaffiliation{A}{NAVER LABS Europe (work started while being at Amazon)}
\icmlaffiliation{B}{Amazon Web Services}

\icmlcorrespondingauthor{David Salinas}{david.salinas@naverlabs.com}
\icmlcorrespondingauthor{Huibin Shen}{huibishe@amazon.com}
\icmlcorrespondingauthor{Valerio Perrone}{vperrone@amazon.com}

\icmlkeywords{Machine Learning, ICML}

\vskip 0.3in
]



\printAffiliationsAndNotice{}  

\begin{abstract}
Bayesian optimization (BO) is a popular methodology to tune the hyperparameters of expensive black-box functions. Traditionally, BO focuses on a single task at a time and is not designed to leverage information from related functions, such as tuning performance objectives of the same algorithm across multiple datasets. In this work, we introduce a novel approach to achieve transfer learning across different \emph{datasets} as well as different \emph{objectives}. The main idea is to regress the mapping from hyperparameter to objective quantiles with a semi-parametric Gaussian Copula distribution, which provides robustness against different scales or outliers that can occur in different tasks. We introduce two methods to leverage this mapping: a Thompson sampling strategy as well as a Gaussian Copula process using such quantile estimate as a prior. We show that these strategies can combine the estimation of multiple objectives such as latency and accuracy, steering the hyperparameters optimization toward faster predictions for the same level of accuracy. Extensive experiments demonstrate significant improvements over state-of-the-art methods for both hyperparameter optimization and neural architecture search.
\end{abstract}

\section{Introduction}
Tuning complex machine learning models such as deep neural networks can be daunting. Object detection or language understanding models often rely on deep neural networks with many tunable hyperparameters, and automatic hyperparameter optimization (HPO) techniques such as Bayesian optimization (BO) are critical to find good hyperparameters in short time. BO addresses the black-box optimization problem by placing a probabilistic model, typically a Gaussian process (GP), on the function to minimize; then, the hyperparameters to evaluate next are determined through an acquisition function that trades off exploration and exploitation.
Interest in BO has been originally motivated by speeding up HPO pipelines where the function to optimize usually takes hours to evaluate or even thousand of GPU days in total in the case of neural architecture search (NAS) \cite{NAS,Real2018}. 
While traditional BO focuses on each problem in isolation, recent years have seen a surge of interest in \emph{transfer learning} for HPO. The key idea is to exploit evaluations from previous, related \emph{tasks} (e.g., the same neural network tuned on multiple datasets) to further speed up the hyperparameter search. 

A key challenge for  joint models is that different black-boxes exhibit heterogeneous scale and noise levels \cite{Bardenet2013,Yogatama2014,Wistuba2018,Feurer2018}.
The straightforward approach of standardizing outputs \cite{Yogatama2014} only works for tasks with normally distributed observations and no outliers. While rank estimation can be used to alleviate scale issues, the difficulty of feeding back rank information to GPs leads to restricting assumptions. For instance, \citet{Bardenet2013} does not model rank estimation uncertainty, while \citet{Feurer2018} uses independent GPs removing the adaptivity of the GP to the current task. 

This paper shows how semi-parametric Gaussian Copulas effectively handle heterogeneous scales across tasks, giving rise to several algorithmic instantiations for hyperparameter transfer learning. Our key contributions are as follows:
\begin{itemize}	
	\item We propose using Gaussian Copulas instead of standardization to map observations from different tasks to comparable distributions;
	\item Two novel methods leveraging this finding, namely a Thompson sampling and Gaussian Copula process combined with a joint parametric prior;
	\item An extensive empirical study demonstrating substantial improvements over state-of-the-art transfer learning methods on real-world datasets, including on neural architecture search (NAS);
	\item A simple extension that scalarizes Gaussian Copula objectives to achieve multi-objective Bayesian optimization.
\end{itemize}

\section{Related work} \label{sec:related_work}
A variety of methods have been developed to induce transfer learning in HPO. The most common approach is to model tasks jointly or via a conditional independence structure, which has been been explored through multi-output GPs~\cite{Swersky2013}, weighted combination of GPs~\cite{Schilling2016,Wistuba2018,Feurer2018}, and neural networks, either fully Bayesian~\cite{Springenberg2016} or hybrid ~\cite{Snoek2015, Perrone2018, Law2018}. A different line of research has focused on the setting where tasks come over time as a sequence and the models need to be updated online as new problems accrue. A way to achieve this is to fit a sequence of surrogate models to the residuals relative to predictions of the previously fitted model~\cite{Golovin2017, Poloczek2016}. Specifically, the GP over the new task is centered on the predictive mean of the previously learned GP. Finally, rather than fitting a surrogate model to all past data, some transfer can be achieved by warm-starting BO with the solutions to the previous BO problems~\cite{Feurer2015, Wistuba2015a}. 

Some methods have instead focused on the search-space level, aiming to prune it to focus on regions of the hyperparameter space where good configurations are likely to lie. An example is \citet{Wistuba2015}, where related tasks are used to learn a promising search space during HPO, defining task similarity in terms of the distance of the respective dataset meta-features. An alternative was proposed in \citet{Perrone2019}, where a restricted search space in the form of a low-volume hyper-rectangle or hyper-ellipsoid is learned from the optimal hyperparameters of related tasks. Gaussian Copula Process (GCP)~\cite{CopulaProcess} can also be used to alleviate scale issues on a single task at the extra cost of estimating the CDF of the data. Using GCP for HPO was proposed in~\citet{anderson2017sample} to handle potentially non-Gaussian data, albeit only considering non-parametric homoskedastic priors for the single task and single objective case.

\section{Gaussian Copula regression}
\label{sec:copula_regression}
For each task $j$ denote with $f^j: \R^\numparameters \to \R$ the error function one wishes to minimize, and with $\mathcal{D} = \{x_i, y_i\}_{i=1}^{N}$ the evaluations available for an arbitrary task with $y_i = f(x_i)$. Given the evaluations on $\numdataset$ tasks 
$$\mathcal{D}^\numdataset = \bigcup_{j=1}^{\numdataset }  \{x_i^j, y_i^j\}_{i=1}^{\numevaluations_j},$$ we are interested in speeding up the optimization of an arbitrary new task $f$, namely in finding $\argmin_{x\in \R^\numparameters}{f(x)}$ in the least number of evaluations. We assume the functions $\{f^j\}_{j=1}^M$ and $f$ to be \emph{related}, such as the error function of the same algorithm over several datasets. In the following, we refer to \emph{task} as the problem of tuning a given algorithm on a dataset, with different datasets corresponding to related tasks.

 One possible approach to speed up the optimization of $f$ is to build a surrogate model $\hat{f}(x)$. While using a Gaussian process is possible, scaling this approach to the large number of evaluations available in a transfer learning setting is challenging. Instead, we propose fitting a parametric estimate of the distribution of $\hat{f}_\theta(x)$ which can be later used in HPO strategies as a prior of a Gaussian Copula Process. A key requirement here is to learn a joint model, namely we would like to find $\theta$ which fits well all observed tasks $f^j$. We show how this can be achieved with a semi-parametric Gaussian Copula in two steps. First, all evaluations are mapped to quantiles with the empirical CDF. Then, we fit a parametric Gaussian distribution on quantiles mapped through the Gaussian inverse CDF.

\begin{figure}
\center
\center
\def\svgwidth{0.4\textwidth}
\begin{tiny}
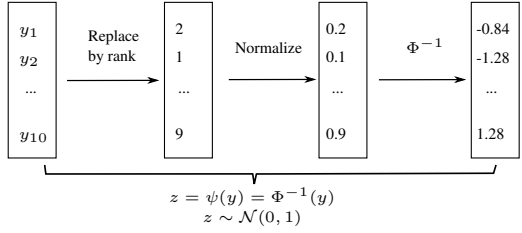
\end{tiny}
\caption{Illustration of the steps to compute of $z = \psi(y)$. The empirical CDF is computed on each task individually and truncated with Winsorized cut-off.
\label{fig:psi-computation}
}
\end{figure}

We make the modeling assumption that every $y_i$ comes from the same distribution for an arbitrary task. The probability integral transform results in $u_i = F(y_i)$, where $F$ is the cumulative distribution function of $y$. The CDF of $(u_1, \dots, u_N)$ is modeled with a Gaussian Copula:
$$C(u_1, \dots, u_N) = \phi_{\mu, \Sigma}(\Phi^{-1}(F(y_1)), \dots, \Phi^{-1}(F(y_{N}))),$$ 
where $\Phi$ is the standard normal CDF and $\phi_{\mu, \Sigma}$ is the CDF of a normal distribution parametrized by $\mu$ and $\Sigma$. Assuming $F$ to be invertible, we define the change of variable $z = \Phi^{-1} \circ F (y) = \psi(y)$ and $g = \psi \circ f$, see Figure \ref{fig:psi-computation}. 
Note that if $z$ is regressed perfectly against $x$, then finding the minimum of $f$ is trivial as a parameter $x$ minimizing $\psi(f(x))$ also minimizes $f(x)$ since $\psi$ is monotonically increasing.

The conditional distribution of $P(z|x)$ is estimated with a Gaussian distribution whose mean and variance are two deterministic parametric functions given by
\begin{align*} 
P(z| x)& \sim \N(\priormu(x), \priorsigma^2(x)) \\ &= \N(w_\mu^T h_{w_h}(x) + b_\mu, (\Psi(w_\sigma^T h_{w_h}(x) + b_\sigma))^2),
\end{align*}
where $h_{w_h}(x) \in \R^d$ is the output of a multi-layer perceptron (MLP), ${w_h}$, $w_\mu\in \R^{d}, b_\mu\in \R, w_\sigma\in \R^{d}, b_\sigma\in \R$ are free parameters and $\Psi(t) = \log(1 + \exp{t})$ is an activation mapping to positive numbers.
The parameters $\mlpparameters=\{{w_h}, w_\mu, b_\mu, w_\sigma, b_\sigma\}$ are learned by minimizing the Gaussian negative log-likelihood on the available evaluations $\mathcal{D^\numdataset} = \bigcup_{1 \leq j \leq \numdataset }  \{x_i^j, z_i^j\}_{i=1}^{N_j}$, e.g., by minimizing with SGD

\begin{equation}
\sum_{(x, z) \in \mathcal{D^\numdataset}}
\frac{1}{2}\log{2\pi\sigma_{\theta}(x)^2} + \frac{1}{2}\left(\frac{z - \mu_{\theta}(x)}{\sigma_{\theta}(x)}\right)^2
\label{eq:quantile-regression-loss}
\end{equation}

\begin{figure*}
\center
\def\svgwidth{0.9\textwidth}
\begin{tiny}
\begingroup%
  \makeatletter%
  \providecommand\color[2][]{%
    \errmessage{(Inkscape) Color is used for the text in Inkscape, but the package 'color.sty' is not loaded}%
    \renewcommand\color[2][]{}%
  }%
  \providecommand\transparent[1]{%
    \errmessage{(Inkscape) Transparency is used (non-zero) for the text in Inkscape, but the package 'transparent.sty' is not loaded}%
    \renewcommand\transparent[1]{}%
  }%
  \providecommand\rotatebox[2]{#2}%
  \newcommand*\fsize{\dimexpr\f@size pt\relax}%
  \newcommand*\lineheight[1]{\fontsize{\fsize}{#1\fsize}\selectfont}%
  \ifx\svgwidth\undefined%
    \setlength{\unitlength}{1036.80002768bp}%
    \ifx\svgscale\undefined%
      \relax%
    \else%
      \setlength{\unitlength}{\unitlength * \real{\svgscale}}%
    \fi%
  \else%
    \setlength{\unitlength}{\svgwidth}%
  \fi%
  \global\let\svgwidth\undefined%
  \global\let\svgscale\undefined%
  \makeatother%
  \begin{picture}(1,0.27777777)%
    \lineheight{1}%
    \setlength\tabcolsep{0pt}%
    \put(0,0){\includegraphics[width=\unitlength,page=1]{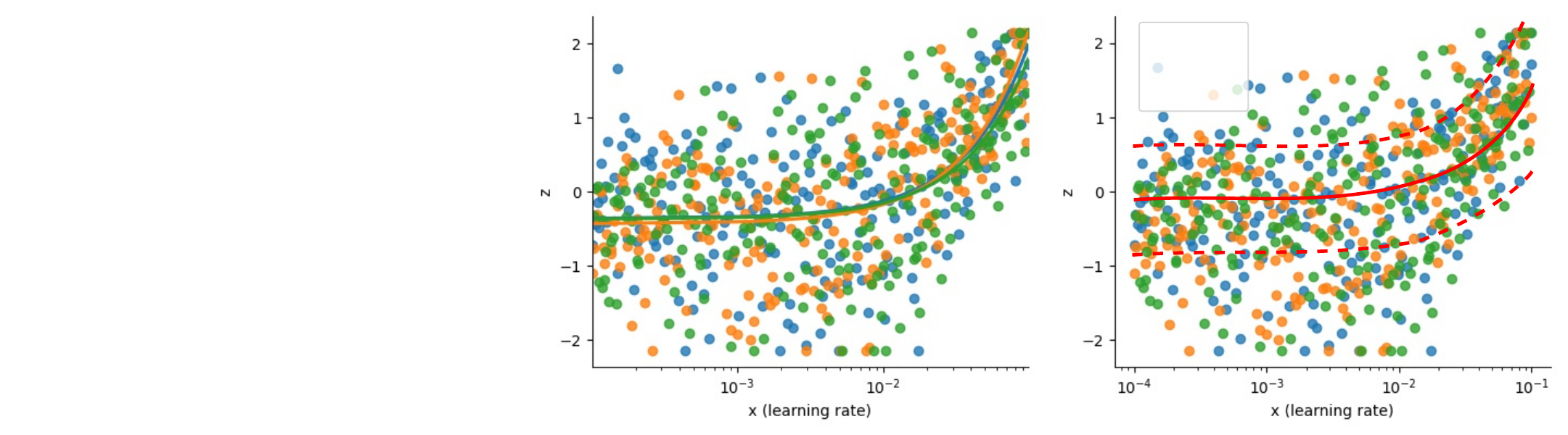}}%
    \put(0.74613645,0.24602262){\color[rgb]{1,0,0}\makebox(0,0)[lt]{\lineheight{1.25}\smash{\begin{tabular}[t]{l}$\mu_\theta(x)$\end{tabular}}}}%
    \put(0.74641254,0.221667){\color[rgb]{1,0,0}\makebox(0,0)[lt]{\lineheight{1.25}\smash{\begin{tabular}[t]{l}$\sigma_\theta(x)$\end{tabular}}}}%
    \put(0,0){\includegraphics[width=\unitlength,page=2]{illustration.pdf}}%
  \end{picture}%
\endgroup%

\end{tiny}
\caption{
Plot of $x$ (single learning rate in this example) against the blackbox error $y$ in log-space (left) and $z = \Phi^{-1}~\circ~F(y)$ (middle and right) where $F$ is approximated with the empirical CDF of every task.
The middle plot shows a running mean fitted for every task individually. 
The last plot pictures the predictive distribution of a parametric prior given by an MLP $\mu_\theta(x), \sigma_\theta(x)$. Note that since $\theta$ is tied across all tasks, the predictions are the same for all tasks which can only work when distributions are similar across tasks.
\label{fig:prior-illustration}
}
\end{figure*}

Let us make two remarks. First, the parameters $\theta$ are \emph{independent} of the task chosen so that minimizing Eq.~(\ref{eq:quantile-regression-loss}) gives a joint model over all available tasks, the hope being it can generalize to new tasks, see Figure \ref{fig:prior-illustration}. Second, each task is assumed to have the same number of observations. If this is not the case, each term in Eq.~(\ref{eq:quantile-regression-loss}) can be weighted inversely to the number of task observations.

The transformation $\psi$ requires $F$, which needs to be estimated. Rather than using a parametric or density estimation approach, we use the empirical CDF $\ECDF(t) = \frac{1}{N} \sum_{i=1}^{N} \mathds{1}_{y_i \leq t}$. While it has the advantage of being non-parametric, this estimator leads to infinite value when evaluating $\psi$ at the minimum or maximum of $y$. To avoid this issue, we use the Winsorized cut-off estimator
$$
F(t) \approx
\begin{cases}
\delta_{N}  & \text{if } \ECDF(t) < \delta_{N}\\
\ECDF(t) & \text{if }  \delta_{N} \leq \ECDF(t) \leq 1 - \delta_{N}\\
1 - \delta_{N} & \text{if } \ECDF(t) > 1 - \delta_{N} \\
\end{cases}
$$
where ${N}$ is the number of observations of $y$ and choosing $\delta_{N} = \frac{1}{4{N}^{1/4}\sqrt{\pi \log {N}}}$ strikes a bias-variance trade-off \cite{Liu2009}. This approach is semi-parametric in that the CDF is estimated with a non-parametric estimator and the Gaussian Copula is estimated with a parametric approach. 

The benefit of using a non-parametric estimator for the CDF is that it allows us to map the observations of each task to comparable distributions.
Indeed, each $u^j = F^j(y^j)$ is uniformly distributed by the probability integral transform property. Since $z^j = \Phi^{-1}(u_j)$, all tasks $j$ are normally distributed, namely $z^j\sim \N(0, 1)$ for all $j$.
 This is critical to allow the joint learning of the parametric estimates $\priormu$ and $\priorsigma$, which share their parameters $\mlpparameters$ across all tasks.

Another advantage of this view is that one can easily assess whether transfer learning helps. Indeed, a constant predictor $\hat{z} = 0$ yields a RMSE of $1$ as

\begin{align*} 
\text{RMSE}(\hat{z})^2 &= \E[(z-\hat{z})^2] \\ 
 & = \E[z^2] - 2~\E[z]\E[\hat{z}] + \E[\hat{z}^2] \\
 & = 1
\end{align*} 
using the independence of $z$ and $\hat{z}$, and the fact  $z \sim \N(0, 1)$.  
Our experiments will show that one can regress a parametric estimate that has a RMSE lower than $1$. This means that information can be leveraged from the evaluations obtained on related tasks, compared to the result of the constant predictor which would be the best predictor if no information was given (assuming of course absence of overfitting).

\section{Copula based HPO}
\label{sec:copula_hpo}
We now show how the estimator introduced in the previous section can be leveraged to design two novel HPO strategies. We first introduce Copula Thompson sampling (CTS), a simple method to exploit information from related tasks. We then build on it to develop Gaussian Copula Process, which can additionally adapt to the new task.

\subsection{Copula Thompson sampling}
\label{sec:CTS}

Given the predictive distribution $P(z|x) \sim \N(\priormu(x), \priorsigma^2(x))$, we can derive a Thompson sampling strategy \cite{thompson} exploiting knowledge from previous tasks. Given $N$ candidate hyperparameter configurations $x_1, \dots, x_N$, we sample from each predictive distribution $\tilde{z_i} \sim \N(\priormu(x_i), \priorsigma^2(x_i))$ and then evaluate $f(x_i)$ where $i = \argmin_{i} \tilde{z_i}$. Pseudo-code is given in Algorithm \ref{algo:cts}. 

While this approach can re-use information from previous tasks, it does not exploit the evaluations from the current task as each draw is independent of the observed evaluations. This can become an issue if the new black-box significantly differs from previous tasks. We now show that Gaussian Copula regression can be combined with a GP to learn from previous tasks while also adapting to the current task.

\begin{algorithm}
\caption{Copula Thompson sampling} 
\label{algo:cts} 
\begin{algorithmic}
\STATE Learn the parameters $\mlpparameters$ of $\priormu(x)$ and $\priorsigma(x)$ on hold-out evaluations $\mathcal{D}^\numdataset$ by minimizing \eqref{eq:quantile-regression-loss}.
\WHILE{Has budget} 
	\STATE Sample $N$ candidate hyperparameters $x_1, \dots, x_N$  from the search space 
	\STATE Draw $\tilde{z_i} \sim \N(\priormu(x_i), \priorsigma^2(x_i))$ for $i=1, \dots, N$
	\STATE Evaluate $f(x_i)$ where $i = \argmin_i{\tilde{z_i}}$ 
\ENDWHILE
\end{algorithmic}
\end{algorithm}

\subsection{Gaussian Copula Process}
\label{sec:GCP}
Instead of modeling observations with a GP, we model them as a Gaussian Copula Process (GCP) \cite{CopulaProcess}.
Observations are mapped through the bijection $\psi = \Phi^{-1}~\circ~F$, where we recall that $\Phi$ is the standard normal CDF and that $F$ is the CDF of $y$. As $\psi$ is monotonically increasing and mapping into the line, we can alternatively view this modeling as a warped GP \cite{WarpedGP} with a non-parametric warping. One advantage of this transformation is that $z = \psi(y)$ follows a normal distribution, which may not be the case for $y$. In the specific case of HPO, for instance $y$ may represent accuracy scores in $[0, 1]$ of a classifier where a Gaussian cannot be used. Furthermore, we can use the information gained on other tasks with $\priormu$ and $\priorsigma$ by using them as prior mean and standard deviation. 
To do so, the following residual is modeled with a GP: 
\begin{align*} 
r(x) & = \frac{g(x) - \priormu(x)}{\priorsigma(x)}\\ 
 &\sim \text{GP}(m(x), k(x, x')),
\end{align*}
where $g = \psi\circ f$.
We use a Mat\'ern-5/2 covariance kernel with automatic relevance determination hyperparameters, and optimize the GP hyperparameters by type-II maximum likelihood \cite{Rasmussen2006}. Categorical hyperparameters are handled by one-hot encoding. Fitting the GP gives the predictive distribution of the residual surrogate $$\surrogate{r}(x) \sim \N(\mu_r(x), \sigma_r^2(x)).$$
Because $\priormu$ and $\priorsigma$ are deterministic functions, the predictive distribution of the surrogate $\hat{g}$ is given by
\begin{align*}
\hat{g}(x) &= \hat{r}(x)\priorsigma(x) + \priormu(x)\\
 &\sim \N(\mu_{\hat{g}}(x), \sigma_{\hat{g}}^2(x))\\
 &\sim \N(\mu_r(x)\priorsigma(x) + \priormu(x), (\sigma_r(x)\priorsigma(x))^2).
\end{align*}

Using this predictive distribution, we can select the hyperparameter configuration maximizing the Expected Improvement (EI) \cite{mockus1978application} of $g(x)$. The EI can then be defined in closed form as
\begin{align*}
\text{EI}(x)&=\E[\max(0,g(x_{\min})- \hat{g}(x))]  \\
&=\sigma_{\hat{g}}(x)  (v(x) \Phi(v(x))+\phi(v(x))),
\end{align*}

\begin{algorithm}
\caption{Gaussian Copula process with parametric prior} 
\label{algo:gcp} 
\begin{algorithmic} 
\STATE Learn the parameters $\mlpparameters$ of $\priormu(x)$ and $\priorsigma(x)$ on hold-out evaluations $\mathcal{D}^\numdataset$ by minimizing \eqref{eq:quantile-regression-loss}.
\STATE Sample an initial set of evaluations $\mathcal{D} = \{(x_i, f(x_i))\}_{i=1}^{N_0}$ via \CTS{}.
\WHILE{Has budget} 	
	\STATE Estimate CDF $F$ on the current task observations $\{f(x_i)\}$ to obtain $\psi$
	\STATE Fit the GP surrogate $\hat{r}$ to the observations $\{(x, \frac{\psi(y) - \priormu(x)}{\priorsigma(x)}) ~|~(x, y) \in \mathcal{D} \}$
   \STATE Sample $N$ candidate hyperparameters $x_1, \dots, x_N$   from the search space 
	\STATE Compute the hyperparameter maximizing EI $x^* = \argmax \text{EI}$
	\STATE Evaluate $y^* = f(x^*)$ and  update 	$\mathcal{D} = \mathcal{D} \cup \{(x^*, y^*)\}$.
\ENDWHILE
\end{algorithmic}
\end{algorithm}

where $v(x) := \frac{\mu_{\hat{g}}(x)  - g(x_{\min})}{\sigma_{\hat{g}}(x)}$, and $\Phi, \phi$ denote the CDF and PDF of the standard normal, respectively. When no observations are available, the empirical CDF $\ECDF$ is not defined. Therefore, we warm-start the optimization on the new task by sampling a set of $N_0 = 5$ hyperparameter configurations via Thompson sampling, as described above. Pseudo-code is given in Algorithm~\ref{algo:gcp}.

\subsection{Computational complexity}
We assume that the tasks in $\mathcal{D}^\numdataset$ contain $n$ observations each, so that we have $Mn$ offline evaluations in total, and that $N$ evaluations have been queried for the new task. Fitting a GP on all tasks costs $O((Mn + N)^3)$, which prevents from using  exact approaches when many offline evaluations are available. As finding parameters of the parametric prior takes $O(Mn)$, running CTS costs $O(Mn + N)$ and running GCP with a parametric prior costs $O(Mn + N^3)$. One benefit of using a parametric prior is to avoid the cubical complexity in the number of offline evaluations. The next section demonstrates that it also improves accuracy compared to state-of-the-art HPO and transfer learning methods.

\section{Experiments}
\label{sec:experiments}
We consider three algorithms in the HPO context: \XGBoost{} \cite{XGBoost}, a 2-layer feed-forward neural network (\FCNET{}) \cite{Klein2019}, and the RNN-based time series prediction model proposed in \citet{deepar} (\DeepAR{}). As advocated in \citet{Eggensperger2012} and \citet{Klein2019}, we compute tabular evaluations (log) uniformly beforehand on multiple datasets to compare methods with sufficiently many random repetitions. Each optimization problem is then discrete as we select from a list of precomputed solutions. While we consider hyperparameter spaces with small to moderate dimensions, optimizing the acquisition over a continuous domain is better suited in higher dimensional spaces. In this setting, GCP is readily applicable, for instance, by optimizing EI with LBFGS. We run each experiment with 30 random seeds on AWS batch with m4.xlarge instances.

We also run experiments on NAS-Bench-201 \cite{nasbench201}. In this benchmark, all possible 15625 configurations of a specific cell search space were evaluated on 3 datasets. Each architecture can be represented as 6 categorical variables, each containing 5 different types of connections. As in \citet{nasbench201}, every model gets as input $x$ the concatenation of the 6 one-hot vectors, resulting in $x\in \R^{30}$. Optimizers are allowed to query the black-box for 70 observations, which corresponds to roughly 12000 seconds in total. More details on each black-box can be found in Table~\ref{tab:task-def}, including the number of hyperparameters (HPs) for each problem. The list of the datasets is in the appendix.

The MLP $h_{w_h}(x)$ used to regress $\priormu$ and $\priorsigma$ has 3 layers with 50 nodes, a dropout rate of 0.1 after each hidden layer and relu activation functions. The learning rate is set to 0.01, and ADAM is run over 1000 gradient updates three times, lowering the learning rate by 5 each time with a batch size of 64. In practice, optimizing the MLP take less than two minutes on a laptop.

\begin{table}
\center
\scriptsize
\begin{tabular}{lrrrr}
\toprule
tasks & \# datasets & \# HPs & \# evaluations  & objective   \\
\midrule
DeepAR & 10 & 6 & 2420 & quantile loss \\
FCNET & 4 & 9 & 248832 & MSE \\
XGBoost & 9 & 9 & 45000 & 1-AUC \\
NAS-Bench-201 & 3 & 6 & 46875 & accuracy, runtime \\
\bottomrule
\end{tabular}
\caption{Statistics of the black-boxes considered. \label{tab:task-def}}
\end{table}

\paragraph{Baselines.}
We compare against random search (\RS{}) and GP-based BO (\GP{}), the two most popular HPO methods, as well as four transfer learning baselines. The first one is warm-start GP (\WSGP{}) \cite{Feurer2015}, which uses the best evaluation from all related tasks to warm-start the GP after standardizing objectives values for each dataset. The second one is \AutoGP{} ~\cite{Perrone2019}, which transfers information by fitting a bounding box that contains the best hyperparameters from each previous task, and runs a GP-based BO in the learned search space. The third one is \SGPT{} \cite{Wistuba2018}, which combines $\numdataset+1$ GPs with a specific acquisition function using rank-matching-based weighting to transfer information across related tasks. The last baseline is \ABLR{} \cite{Perrone2018}, a multi-task model consisting of a shared-across-task neural network with a per-task Bayesian linear regression layer on top.

For NAS, we also compare with the 4 best methods out of the 10 considered in \citet{nasbench201}: \RS{}, REINFORCE \cite{reinforce}, BOHB \cite{bohb} and REA \cite{REA}.\footnote{These baselines are not evaluated on other black-boxes as they are either designed for NAS or require multi-fidelity information that is not available for all black-boxes.} The transfer learning capabilities of each method are evaluated in a leave-one-task-out setting: one dataset is sequentially left out to assess how much transfer can be achieved from the other datasets, and overall results are aggregated.

\begin{figure*}
\center
 \includegraphics[width=0.75\textwidth]{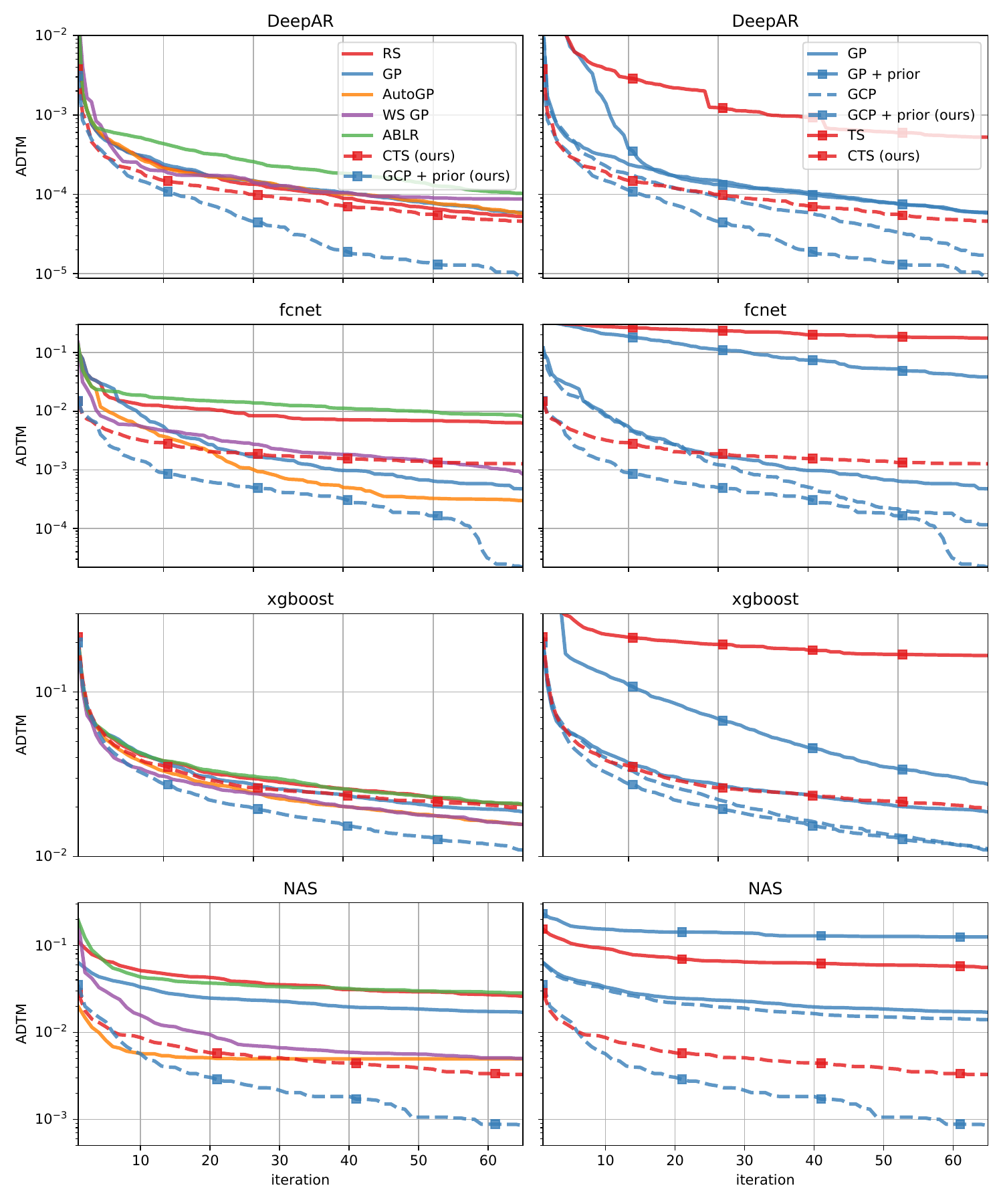}
\caption{ADTM over iterations for baselines (left) and ablation variants (right). Methods using a parametric prior are depicted with a square and methods using Gaussian Copula are represented by a dashed line.
\label{fig:ADTM-average}
}
\end{figure*}

\begin{figure*}
\center
\includegraphics[width=0.99\textwidth]{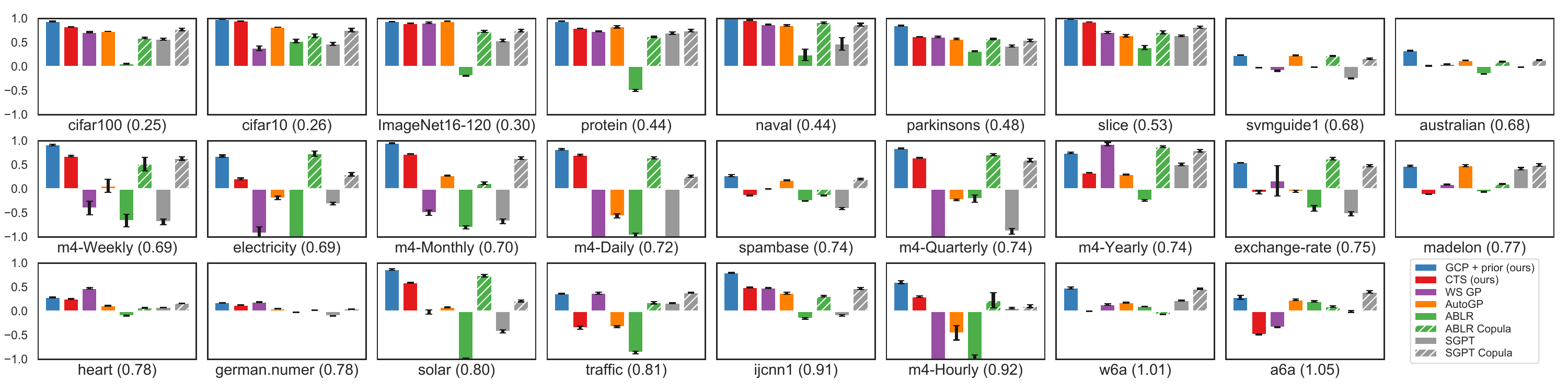}
\caption{DTM improvement over \RS{} (higher is better) for datasets of \emph{all} black-boxes. Datasets are sorted by increasing RMSE error, which is indicated in parenthesis. Transfer learning difficulty is the lowest for the top-left task and increases when going to the right. Most transfer learning methods struggle to keep good performance as the difficulty of transfer learning increases.
\label{fig:ADTM-per-dataset}
}
\end{figure*}

\begin{figure*}
\center
\includegraphics[width=0.32\textwidth]{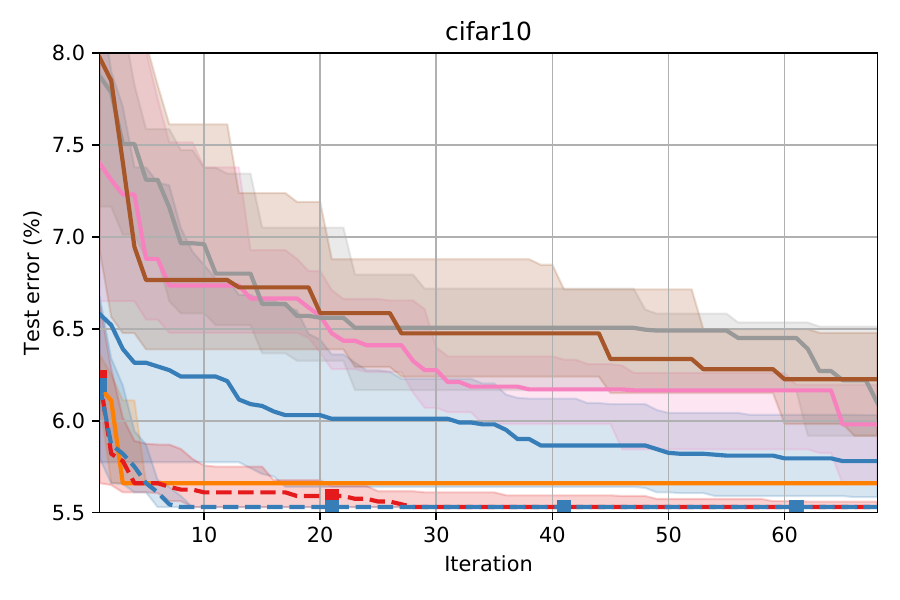} 
\includegraphics[width=0.32\textwidth]{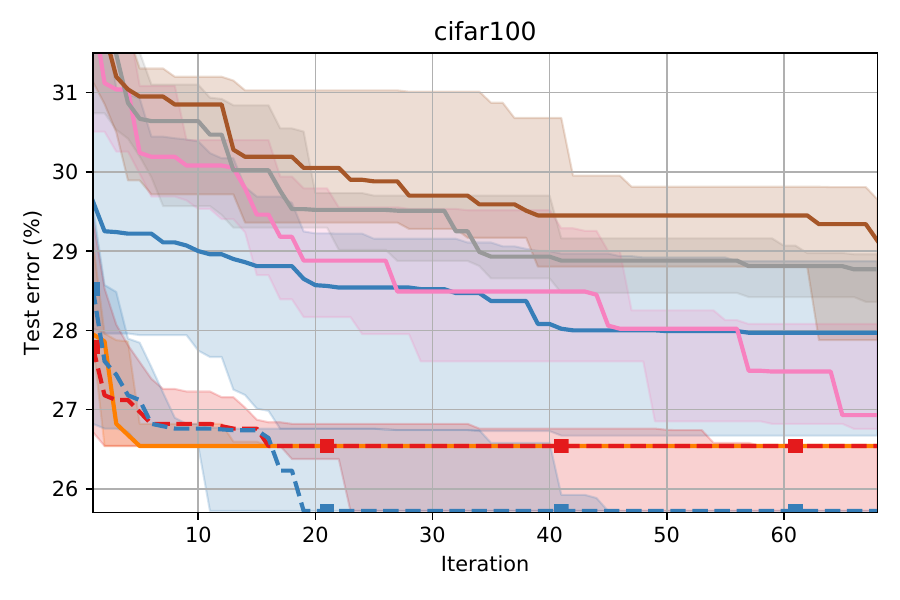}
\includegraphics[width=0.32\textwidth]{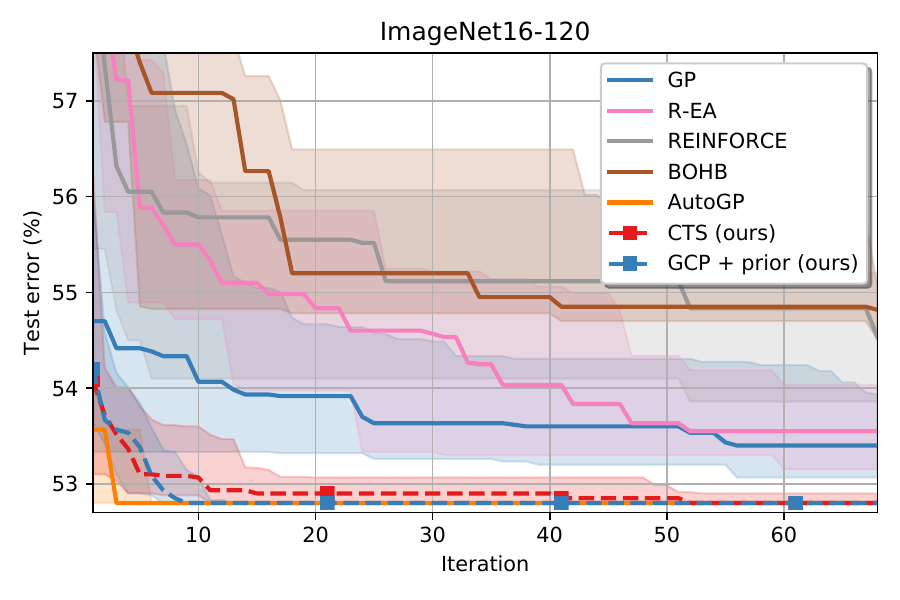}
\caption{Performance comparison on NAS on the 3 NAS-Bench-201 datasets, with shaded areas representing 80\% confidence intervals.\label{fig:nas}}
\end{figure*}

\paragraph{Ablation study.}

In addition to comparing to baselines, we perform an ablation to measure the benefits of 1) using a Gaussian Copula Process rather than sampling independently with Thompson sampling, 2) applying Gaussian Copula to normalize the data using $\psi$, and 3) using a parametric prior. 
When $\psi$ is not used, data is instead standardized with 
$$z=\standardize(y) = \frac{y - \E(y)}{\sqrt{\mathbf{V}(y)}}.$$

Specifically, we evaluate the following variants of our model:
\begin{itemize}
\item \CTS{}: The CTS model with a parametric prior described in Section~\ref{sec:GCP}.
\item \TS{}: The model with a parametric prior described in Section~\ref{sec:GCP} where data is standardized with $\standardize$ instead of $\psi$.
\item \GCPprior{}: The GCP model with a parametric prior described in Section~\ref{sec:GCP}.
\item \GCP{}: A Gaussian Copula Process with a standard prior.
\item \GPprior{}: A Gaussian Process where a parametric prior is estimated but data is standardized with $\standardize$ instead of using $\psi$.
\end{itemize}

\paragraph{Average distance to the minimum.}
To be able to aggregate scores over tasks, we follow the approach from \citet{Wistuba2018} and measure performance on each task in terms of the normalized distance to the global minimum. This is defined as
$$\DTM_\text{opt}^j(t) = \frac{y_{\text{opt}}^j(t) - y_{\text{min}}^j}{y_{\text{max}}^j - y_{\text{min}}^j},$$
where $y_{\text{opt}}^j(t)$ denotes the best performance, averaged over replicates of an optimizer after $t$ iterations on task $j$, while $y_{\text{min}}^j$ and $y_{\text{max}}^j$ respectively denote the minimum and maximum objective computed across all offline evaluations available for task $j$. This score is in $[0, 1]$, making performance more comparable across datasets. The average DTM across tasks is defined as $\ADTM_{\text{opt}}(t) = \frac{1}{\numdataset}\sum_{j=1}^\numdataset\DTM_\text{opt}^j(t)$. 

Figure \ref{fig:ADTM-average} illustrates the performance of competing methods over time for each black-box in terms of ADTM. We report the mean ADTM across all seeds, noting that in this transfer learning setting standard deviation would emphasize the variance coming from the different meta-datasets. Additionally, Table \ref{tab:ablation} reports the average improvement over \RS{}, defined as the average across datasets of $\frac{1}{T}\sum_{t=1}^T\frac{\DTM_\text{RS}^j(t) - \DTM_\text{opt}^j(t)}{\DTM_\text{RS}^j(t)} \in \mathopen]-\infty,1\mathclose]$. This shows how much each algorithm improves over \RS{}, whose performance indicates the complexity of the tuning problem. Figure \ref{fig:ADTM-per-dataset} shows the improvement over random search on each dataset.

\subsection{Results Discussion}

Figure \ref{fig:ADTM-average} and Table \ref{tab:ablation} show that the Copula approach gives consistent improvement over both \GP{} and TS. In particular, \GCP{} is a strong baseline, which is expected as the modeled data after $\psi$ is Gaussian as opposed to a standard \GP{}. Critically, using a parametric prior is only beneficial in combination with Gaussian Copula as evidenced by the very poor performance of \TS{} and \GPprior{}.
This issue also affects \ABLR{} and \SGPT{}, which are unable to consistently outperform \GP{} even though they leverage additional information from other tasks while \AutoGP{} and \WSGP{} are less affected as they only use the best hyperparameters evaluation of each task. In addition, Figure \ref{fig:ADTM-per-dataset} and Table \ref{tab:ablation} report results when using Gaussian Copula in combination with these baselines (e.g., using $\psi$ instead of $\Delta$ to normalize outputs). The quality of these methods is dramatically improved, showing how they are negatively affected by heterogeneous scales and non-normality.

While being able to transfer information from other datasets, \CTS{} is unable to benefit from observations of the current task and is outperformed by other baselines given sufficiently many observations, especially on \DeepAR{} and \XGBoost{}. On these black-boxes, we observe modest performance for the other transfer learning baselines, which we believe is due to the lower correlation of hyperparameter performance between tasks. To investigate this further, Figure \ref{fig:ADTM-per-dataset} shows the average improvement over random search computed separately on each dataset and sorted by the prior RMSE computed on the current task\footnote{Observations from the current task are only used to report RMSE but are not used when fitting our model.} with $\sqrt{\frac{1}{N}\sum_{i=1}^{N} \left(z_i - \mu_\theta(x_i)\right)^2}$. As mentioned in Section \ref{sec:copula_regression}, low RMSE values indicate that the current task is similar to other available tasks and consequently easier for transfer learning methods. 
Both \CTS{} and transfer learning baselines show improvement over \RS{} when the RMSE is low, while the performance of baselines deteriorates for tasks with higher RMSE and is even adversely affected when the test task excessively differs from the held-out datasets. 
On the other hand, being able to benefit both from other tasks and observations of the current task, \GCPprior{} is the best method overall. This can be observed on all black-boxes both at the beginning and at the end of the optimization. The results are summarized in Table \ref{tab:ablation}, which gives the average rank of the 16 methods. Over the 26 datasets, \GCPprior{} is the best method 15 times and the second best 7 times, with an average rank of 1.5.

\paragraph{Comparison with NAS.}
Figure \ref{fig:nas} shows test accuracy over time compared to the 4 best baselines evaluated in \citet{nasbench201}. Interestingly, \GP{} appears to be a satisfactory baseline even though it is rarely evaluated in this context \cite{nasbench101, nasbench201}.
When combined with a prior, both \GCP{} and CTS converge in a fraction of the number of iterations required by the other baselines. The only exception is \AutoGP{}, which \GCPprior{} still outperforms given sufficiently many observations due to its greater ability to adapt to the target task.

\begin{table}
\caption{DTM normalized over random search (higher is better). The best two methods are in bold and the average rank of each method is indicated in parenthesis. \label{tab:ablation}}
\center
\scriptsize
\setlength{\tabcolsep}{4pt}
\begin{tabular}{lr@{\hspace{0.3\tabcolsep}}rr@{\hspace{0.3\tabcolsep}}rr@{\hspace{0.3\tabcolsep}}rr@{\hspace{0.3\tabcolsep}}r}
\toprule
black-box &  \multicolumn{2}{c}{DeepAR} &  \multicolumn{2}{c}{FCNET} &   \multicolumn{2}{c}{XGBoost} &    \multicolumn{2}{c}{NAS} \\
\midrule
RS (baseline)      &     0.00 &(7.1) &     0.00 &(10.8) &    0.00 &(8.2) &   0.00&(12.0) \\
\midrule
TS                 &  -21.02 &(13.0) &  -563.27 &(13.0) &  -6.28 &(12.7) &  -1.30 &(14.3) \\
CTS (ours)         &     0.38 &(4.5) &      \bf{0.83} &(2.5) &    0.02& (7.4) &    \bf{0.88}& (2.7) \\
GP + prior         &   -5.92 &(11.8) &  -166.64 &(12.0) &  -1.70 &(11.1) &  -2.24 &(15.3) \\
GCP                &     0.42 &(4.3) &      0.79 &(4.0) &    \bf{0.31} & (3.1) &    0.45& (7.3) \\
GCP + prior (ours) &     \bf{0.73} &(1.7) & \bf{0.94} &(1.0) &  \bf{0.37} & (1.9) &    \bf{0.94}& (1.3) \\
\midrule
GP                 &    -0.25 &(7.9) &      0.53 &(8.0) &    0.00 &(8.6) &    0.38&(8.7) \\
AutoGP             &    -0.11 &(7.3) &      0.72 &(5.2) &    0.22 &(4.2) &    0.84 &(2.3) \\
WS GP              &    -0.50 &(7.6) &      0.73 &(5.2) &    0.11 &(5.9) &    0.62 &(5.7) \\
ABLR               &   -0.75 &(10.2) &     0.11 &(10.2) &   -0.05 &(9.1) &   0.13 &(10.3) \\
ABLR Copula        &     \bf{0.53} &(3.1) &      0.71 &(5.5) &    0.08 &(7.0) &    0.63 &(5.3) \\
SGPT               &    -0.38& (8.8) &      0.56 &(8.2) &   -0.01 &(8.4) &    0.46& (8.0) \\
SGPT Copula        &     0.44& (3.7) &      0.74 &(5.2) &    0.28 &(3.3) &    0.67 &(5.0) \\
BOHB               &      - & &       - & &     - & &  -0.19& (14.3) \\
R-EA               &      - & &       - & &     - & &   0.19& (10.3) \\
REINFORCE          &      - & &       - & &     - & &  -0.09 &(13.0) \\
\bottomrule
\end{tabular}
\end{table}

\begin{figure*}
\center
\includegraphics[width=0.99\textwidth]{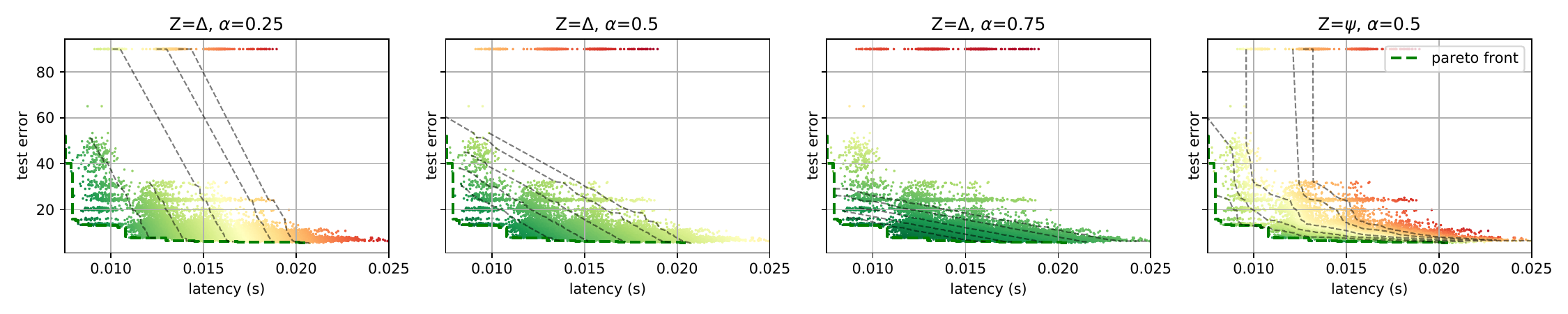}

\caption{Latency and test error for each architecture evaluated on Cifar10. Colors represent values scalarized with $\alpha Z(y_{\text{error}}) + (1-\alpha) Z(y_{\text{latency}})$ for $\alpha \in \{0.25, 0.5, 0.75\}, Z = \standardize$ for the first three plots, and $\alpha = 0.5, Z=\psi$ for the rightmost plot (lowest values are green, highest values are red). 
Level sets are linear for $Z = \standardize$ and no single value of $\alpha$ can approximate the geometry of the Pareto front. In contrast, $\psi$ better approximates the shape of the front, which is followed closely by the top values of the scalarized objective.
\label{fig:scalarization}
}
\end{figure*}

\begin{figure*}
\center
\includegraphics[width=0.99\textwidth]{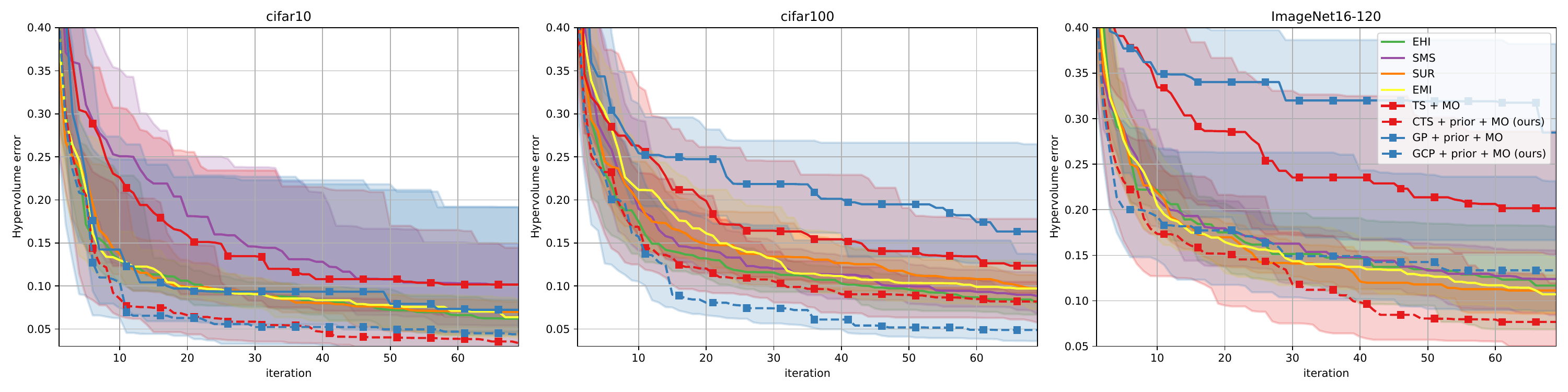}
\caption{
NAS multi-objective experiments. Hypervolume error at each iteration, with shaded areas representing 80\% confidence intervals. 
Methods using a parametric prior are depicted with a square and methods using Gaussian Copula are represented by a dashed line.
\label{fig:hypervolume}}
\end{figure*}

\subsection{Multi-Objective Optimization}

The goal of multi-objective optimization (MO) is to optimize multiple objectives $f_1, \dots, f_\numobjective$ \emph{simultaneously}. This is relevant to many applications including NAS, where the device on which the model is deployed comes with additional memory or latency constraints \cite{Tan2018,Elsken2019}. Typically, no single solution minimizes all objectives at once and one seeks instead Pareto-optimal solutions. A solution $x$ dominates $x'$ if 
$f_l(x) \leq f_l(x')$ for all $l \leq \numobjective$ and there exists $l'\leq \numobjective$ such that $f_{l'}(x) < f_{l'}(x')$. The Pareto front is the set of all Pareto-optimal solutions, defined as points that are not dominated by any other points. 

A simple approach to MO is to scalarize the objective values $y_1, \dots, y_\numobjective$ as $\sum_{l=1}^\numobjective \alpha_l y_l$ and fall back to single-objective minimization. However, combining multiple objectives poses challenges similar to the ones from the transfer learning setting. Objectives typically have different scales and mixing them linearly only allows for convex level sets, which is a poor approximation of the Pareto frontier geometry. We illustrate this behavior in Figure~\ref{fig:scalarization}: no mixture coefficient properly approximates the Pareto front of latency and prediction error on Cifar10. On the other hand, as \citet{Binois2015} observed in the context of Pareto front estimation, averaging the two Gaussian Copula objectives provides a good approximation of the Pareto front.

Motivated by this property, we extend our methods to MO by simply averaging observations from Gaussian Copulas. We compare with \EHI{} \cite{EmmerichEHI}, \SMS{} \cite{PonweiserSMS}, \SUR{} \cite{Picheny2015SUR} and \EMI{} \cite{SvensonEMI}, implemented in GPareto \cite{Binois2019}. \todo{explain methods in related work} The suffix \texttt{+MO} is used to indicate a scalarization of the objective obtained by averaging observations after applying $Z=\psi$ for methods using Copula and $Z=\Delta$ for others. Performance at each BO iteration is evaluated by computing the Pareto hypervolume error, namely the hypervolume difference with the Pareto front. Consistently with the results of the previous section, linear scalarization performs poorly and both GP+prior and TS are strongly outperformed while \GCPprior{} and \CTS{} compete or outperform GPareto baselines. 

\section{Hardware Specification}
We used AWS batch with m4.xlarge instances for most of our experiments. Beside \RS{} whose cost is almost negligible, evaluating an optimizer takes around 5 minutes for a seed. Excluding GPareto and NAS baselines, we then estimate the cost of running our experiments to be $\text{num\_methods} \times \text{num\_seeds} \times \text{num\_datasets} \times \text{optimizer\_time} \approx 12 \times 30 \times 26 \times 300$, which is around 32 days of a single machine.

\section{Conclusions} \label{sec:conclusions}
We introduced a new class of methods to accelerate hyperparameter optimization by exploiting evaluations from previous tasks. The key idea was to leverage a semi-parametric Gaussian Copula prior to account for the different scale and noise levels across tasks. Experiments showed that competing methods are outperformed on both HPO and NAS, and our approach deals with heterogeneous tasks more robustly than a number of transfer learning strategies recently proposed in the literature. Finally, we showed that our framework seamlessly extends to combine multiple metrics, such as test error and latency, in a multi-objective BO framework.

A number of directions for future work are open. First, one could combine our Copula-based HPO strategies with Hyperband-style optimizers~\citep{Li2016}. In addition, one could generalize our approach to deal with settings in which related problems are not limited to the same algorithm run over different datasets. This would allow for different hyperparameter dimensions across tasks, or perform transfer learning across different black-boxes.

\section*{Acknowledgements}

We thank anonymous reviewers as well as Arnaud Sors, Matthias Seeger, Aaron Klein and Michele Donini whose feedback  greatly improved this paper.

\bibliography{quantile_hpo_icml}
\bibliographystyle{icml2020}

\newpage

\appendix

\section{Code}

The code to reproduce the results of the paper will be made available on this github repository \footnote{\url{https://github.com/geoalgo/A-Quantile-based-Approach-for-Hyperparameter-Transfer-Learning/}}. Offline evaluations of blackboxes are already available on this repository \footnote{\url{https://github.com/icdishb/hyperparameter-transfer-learning-evaluations/}}.

\section{Baselines Details}

\paragraph{\WSGP{}.} This method uses the best-performing hyperparameter configuration from each related task to warm start the GP on the target task \citep{Feurer2015}. We also compared to two variants of \WSGP{}: the first-one uses dataset meta-features to detect the most closely related task and warm-starts the GP with the best $s$ evaluations from that task; a second variant takes the best-performing $s$ evaluations from each task, with $s>1$. As both variants were outperformed by taking only the best evaluation from each task (i.e., $s=1$), we show results against this version in the paper.

\paragraph{\ABLR{}.} The same algorithm settings as in \citet{Perrone2018} are used. The shared neural network consists of three fully connected layers, each with 50 units and \texttt{tanh} activation functions. Its weights, as well as the task-specific scale and noise parameters, are learned by optimizing the marginal log-likelihood by L-BFGS. To run BO, ABLR is combined with the EI acquisition function. We restrict the total number of evaluations for transfer learning to be around $2,000$ so that it is computationally feasible to train with L-BFGS.

\paragraph{\SGPT{}.} The method fits independent GPs on each related task and weights tasks based on rank matching between the objective values from the target task and the predictions of these GPs on the evaluated hyperparameter configurations \citep{Wistuba2018}. 
The weights are further included in the acquisition function to scale the predictive improvement on every relevant task. Due to the cubical scaling of GPs, we subsampled $1,000$ hyperparameter evaluations from each related task. We found the results to be very sensitive to the choice of $\rho$, that is the bandwidth of the ranking-based distance defined in Eq.\ (42) of \citet{Wistuba2018}. We report all SGPT results with $\rho=0.01$, which gives the best overall performance across tasks among $\rho\in\{1.0, 0.1, 0.01\}$. 

\paragraph{R-EA.} As in \citet{nasbench201}, the initial population size is 10, the number of cycles is set to infinity, and the sample size is set to 3. 
	
\paragraph{REINFORCE.} As in \citet{nasbench201}, the architecture encoding is optimized with ADAM. The learning rate is set to 0.001 and the momentum for exponential moving average is set to 0.9.
	
\paragraph{BOHB.} As in \citet{nasbench201}, the number of samples for the acquisition function is set to 4, the random fraction is set to 0, the minimum bandwidth is set to 0.3 and the bandwidth factor to 3. 
	
\paragraph{AutoGP.} This method transfers information by learning a compact search space from other tasks \cite{Perrone2019}. First, a bounding box containing the best hyperparameter configuration from each other task is fit to obtain a smaller search space, which is defined by the learned coordinate-wise lower and upper bounds. Then, standard random search (\RS{}) or GP-based BO (\GP) is run in the learned search space. This method comes with no hyperparameters.

\paragraph{GPareto.} We use the four different criteria, namely EHI, SMS, SUR and EMI, from GPareto \cite{Binois2019}. When considering the new candidate with EI, we select the best possible option over the known grid of candidates. Importantly, for GPareto this search becomes prohibitively slow so we maximize EI over a random sample of 2000 candidates out of 15625. To compute the Hypervolume error, the maximum of latency and error is used as a reference point.

\section{Look-up Tables}

Table \ref{tab:parameter-range} describes the hyperparameters considered for each tuning problem.
For DeepAR and XGBoost, evaluations were obtained by sampling hyperparameters (log) uniformly at random from their search space. For FCNET and NAS, all possible configurations were evaluated. 

For DeepAR, we used the following 10 public datasets from GluonTS \cite{GluonTS}: \{electricity, traffic, solar, exchange-rate, m4-Hourly, m4-Daily, m4-Weekly, m4-Montly, m4-Quarterly, m4-Yearly\}\footnote[1]{Datasets available at \url{https://github.com/awslabs/gluon-ts/blob/master/src/gluonts/dataset/repository/datasets.py}.}. The method was evaluated with the Sagemaker version \cite{januschowski2018now}.
For FCNET and NAS, we used evaluations from \citet{Klein2019} and \citet{nasbench201} which contains evaluations on \{parkinson, protein, naval, slice\} for FCNET
 and \{cifar10, cifar100, Imagenet16\} for NAS. For XGBoost, we used the tree boosting implementation from \citet{XGBoost} and evaluated $5,000$ hyperparameter configurations against $9$ LIBSVM binary classification datasets, namely \{a6a, australian, german.numer, heart, ijcnn1, madelon, spambase, svmguide1, w6a\}.\footnote[2]{Datasets from \url{https://www.csie.ntu.edu.tw/~cjlin/libsvmtools/datasets/binary.html}.}

\begin{table*}[h]
\caption{Search spaces description for each blackbox.
 \label{tab:parameter-range}}
\scriptsize
\center
\begin{tabular}{lrrr}
\toprule
tasks &  hyperparameter & search space & scale\\
\midrule
\multirow{6}{*}{\DeepAR{}} & \# layers & [$1$, $5$] & linear \\
 & \# cells & [$10$, $120$] & linear \\
 & learning rate & [$10^{-4}$, $0.1$] & log10 \\
 & dropout rate & [$10^{-2}$, $0.5$] & log10 \\
 & context\_length\_ratio & [$10^{-1}$, $4$] & log10 \\
 & \# bathes per epoch & [$10$, $10^4$] & log10 \\
\hline
\multirow{9}{*}{\XGBoost{}} & num\_round & [$2$, $2^9$] &  log2\\
 & eta & [$0$, $1$] & linear\\
 & gamma & [$2^{-20}$, $2^6$] & log2\\
 & min\_child\_weight & [$2^{-8}, 2^6$] & log2\\
 & max\_depth & [$2$, $2^7$] & log2\\
 & subsample & [$0.5$, $1$] & linear\\
 & colsample\_bytree & [$0.3$, $1$] & linear\\
 & lambda & [$2^{-10}$, $2^8$] & log2\\
 & alpha & [$2^{-20}$, $2^8$] & log2\\
 \hline
 \multirow{9}{*}{\FCNET{}} & initial\_lr & \{$0.001$, $0.005$, $0.01$, $0.05$, $0.1$\} & categorical  \\
 & batch\_size & \{$8$, $16$, $32$, $64$\} &categorical \\
 & lr\_schedule & \{cosine, fix\} & categorical \\
 & activation layer 1 & \{relu, tanh\} & categorical \\
 & activation layer 2 & \{relu, tanh\} & categorical \\
 & size layer 1 & \{$16$, $32$, $64$, $128$, $256$, $512$\} & categorical \\
 & size layer 2 &\{$16$, $32$, $64$, $128$, $256$, $512$\} & categorical \\
 & dropout layer 1 & \{$0.0$, $0.3$, $0.6$\} & categorical \\
 & dropout layer 2 & \{$0.0$, $0.3$, $0.6$\} & categorical \\

 \hline
 \multirow{6}{*}{\NAS{}} & $\text{edge}_1$ & \{zeroize, skip, 1x1 conv, 3x3 conv, 3x3 avg pool\} & categorical \\
 & $\text{edge}_2$ & \{zeroize, skip, 1x1 conv, 3x3 conv, 3x3 avg pool\} & categorical \\
 & $\text{edge}_3$ & \{zeroize, skip, 1x1 conv, 3x3 conv, 3x3 avg pool\} & categorical \\
 & $\text{edge}_4$ & \{zeroize, skip, 1x1 conv, 3x3 conv, 3x3 avg pool\} & categorical \\
 & $\text{edge}_5$ & \{zeroize, skip, 1x1 conv, 3x3 conv, 3x3 avg pool\} & categorical \\
 & $\text{edge}_6$ & \{zeroize, skip, 1x1 conv, 3x3 conv, 3x3 avg pool\} & categorical \\
\bottomrule
\end{tabular}

\end{table*}

Table \ref{tab:parameter-range} describes the hyperparameters considered for each tuning problem.
For DeepAR and XGBoost, evaluations were obtained by sampling hyperparameters (log) uniformly at random from their search space. For FCNET and NAS, all possible configurations were evaluated.

\end{document}